\pdfoutput=1

\documentclass[11pt]{article}

\usepackage[]{acl}

\usepackage{multicol}

\usepackage{times}
\usepackage{latexsym}
\usepackage{tcolorbox}

\usepackage[T1]{fontenc}

\usepackage{array}
\newcolumntype{P}[1]{>{\centering\arraybackslash}p{#1}}

\usepackage[utf8]{inputenc}

\usepackage{microtype}

\usepackage{inconsolata}

\usepackage{bbm}
\usepackage{amsmath}
\usepackage{amssymb}
\DeclareMathOperator*{\argmax}{arg\,max}

\usepackage{pifont}
\newcommand{\xmark}{\ding{55}}%

\usepackage{amsmath,amssymb,amsfonts}
\usepackage{bm}

\usepackage{xcolor,colortbl}

\usepackage{multirow, booktabs}
\usepackage{caption}
\usepackage{subcaption}
\usepackage{makecell}

\usepackage{graphicx}

\newcolumntype{P}[1]{>{\centering\arraybackslash}p{#1}}

\usepackage{pifont}

\title{CUED at ProbSum 2023: Hierarchical Ensemble of Summarization Models}

\author{
\textbf{Potsawee Manakul, Yassir Fathullah, Adian Liusie,}\\\textbf{Vyas Raina, Vatsal Raina, Mark Gales} \\
  ALTA Institute, Engineering Department, University of Cambridge \\
  \texttt{\{pm574,yf286,al826,vr313,vr311,mjfg\}@cam.ac.uk}
  \\}

\begin{document}
\maketitle
\begin{abstract}


In this paper, we consider the challenge of summarizing patients' medical progress notes in a limited data setting. For the Problem List Summarization (shared task 1A) at the BioNLP Workshop 2023, we demonstrate that Clinical-T5 fine-tuned to 765 medical clinic notes outperforms other extractive, abstractive and zero-shot baselines, yielding reasonable baseline systems for medical note summarization. Further, we introduce Hierarchical Ensemble of Summarization Models (HESM), consisting of token-level ensembles of diverse fine-tuned Clinical-T5 models, followed by Minimum Bayes Risk (MBR) decoding. Our HESM approach lead to a considerable summarization performance boost, and when evaluated on held-out challenge data achieved a ROUGE-L of 32.77, which was the best-performing system at the top of the shared task leaderboard.\footnote{Our code is available at \url{https://github.com/potsawee/hierarchical_ensemble_summ}.}

\end{abstract}

\section{Introduction}
Summarization is a common natural language generation (NLG) task with growing recent interest \cite{elkassas2021survey}. The 1A shared task of BioNLP 2023 considers medical problem list summarization \cite{gao-etal-2023-overview}, where patient notes are summarized to assist medical diagnosis applications. There are several challenges faced in designing systems for this task: First, the challenge is low-resource, with only 765 examples available for training/validation. 
Second, a high-stake application with specialized medical terms requires systems that can deal with domain-specific terms and find relevant diagnoses from patient documents.

\begin{figure}[!t]
\centering
\includegraphics[width=\linewidth,keepaspectratio]{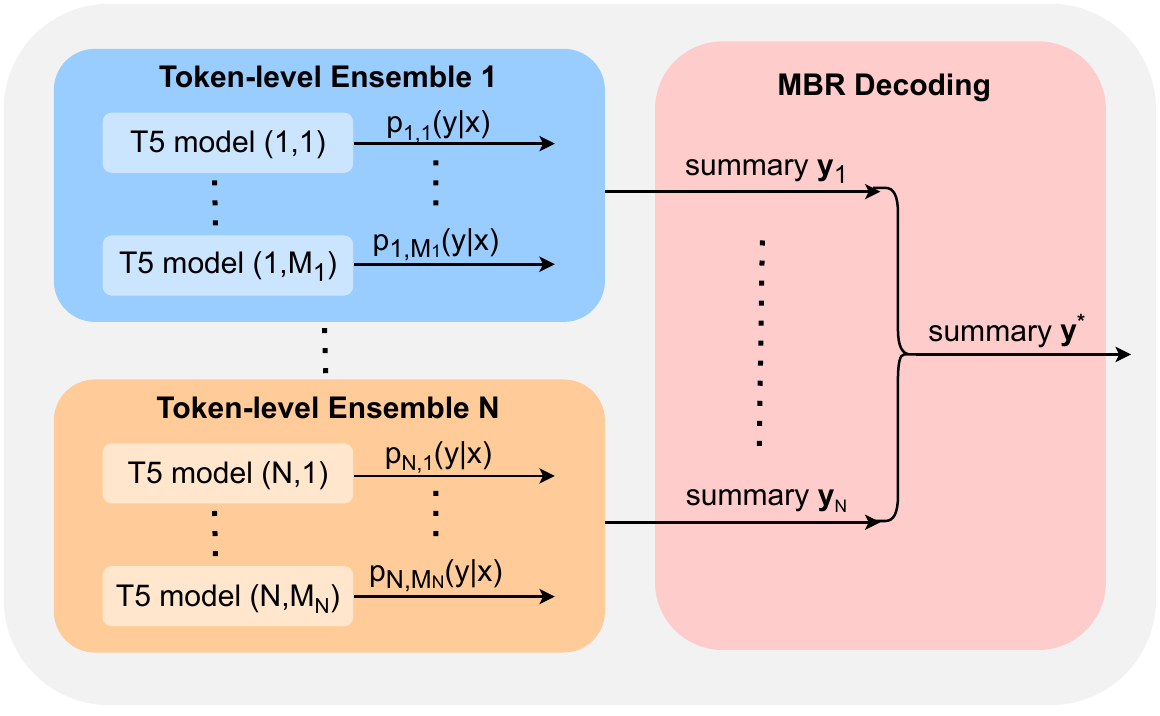}
    \caption{Hierarchical ensemble of summarization models where each individual model is a fine-tuned Clinical-T5.}
    \label{fig:diagram}
\end{figure}

This paper introduces Hierarchical Ensemble of Summarization Models (HESM), an approach that is composed of two sequential ensembling layers, of token-level ensembles, followed by Minimum Bayes Risk (MBR) decoding, as shown in Figure \ref{fig:diagram}. Ensembling methods combine the predictions of various models \cite{fort2020deep} and have been effective in NLG tasks, such as in summarization \cite{manakul2020cued_speech}. For low-resource settings, this method allows outputs to be composed from multiple different ensemble members, which can reduce the influence of noise and eliminate spurious signals, reducing the chance of medically inaccurate summaries. We demonstrate that using HESM with Clinical-T5 models \cite{lehman2023we} leads to systems that have a good grasp of medical knowledge, and that are able to generate outputs that are consistently closer to ground-truth summaries. Our proposed HESM method was submitted to the BioNLP problem list summarization challenge, and achieved the top position of the shared task leaderboard, out of 9 teams.

\section{Background and Related Work}



\subsection{Existing Pre-Trained Language Models}
A common approach for current NLP applications has been the pre-train and fine-tuning paradigm, where pre-trained models are fine-tuned to specific target tasks. The community has open-sourced a variety of pre-trained backbones of different sizes and architectures, including encoder-only such as BERT \cite{devlin-etal-2019-bert}, decoder-only such as GPT-2 \cite{radford2019language}, and encoder-decoder such as BART \cite{lewis-etal-2020-bart} and T5 \cite{raffel2020exploring}. Medical domain versions of language models have also been created, including ClinicalBERT \cite{clinicalbert}, BioBERT \cite{lee2020biobert}, BioMed-RoBERTa \cite{gururangan-etal-2020-dont}, BioGPT \cite{10.1093/bib/bbac409}, and Clinical-T5 \cite{lehman2023we}.  

\subsection{Summarization Methods}
The two main summarization approaches are extractive methods, which select relevant words/phrases present in the input for the summary, and abstractive methods, which can freely generate text (even text that may not be present in the source). Previous work in medical list summarization demonstrated significant gain from adapting BART and T5 (both are abstractive models) to the medical domain and by fine-tuning them for summarization \cite{gao-etal-2022-summarizing}. For long-input summarization, a preliminary stage of selecting the most relevant input sentences before summarization was shown to be effective \cite{manakul-gales-2021-long}. Alternatively, large language models have recently shown success in zero-shot summarization \cite{brown2020language}.


\section{Hierarchical Ensemble of Summarization Models}



When working with a small dataset, individual models are prone to overfit specific aspects of the data due to the limited number of training samples. By training multiple models on the same dataset, each model can potentially capture different aspects of the data that are generalizable and not prone to overfitting. Combining these diverse models together can then create a more robust and accurate prediction model \cite{consensus-decoding}. 


Various approaches can be used to create diverse individual systems. A simple approach is to use different weights' initialization \cite{NIPS2017_9ef2ed4b} for different seeds. Alternatively for pre-trained systems (as considered in this work), one can set different random seeds, which will influence training dropout and stochastic gradient descent batch creation, resulting in variability in the final models' weights. One can also use a form of data \textit{bagging}, where a different subset of the data is used to train each model \cite{galar2011review}. For example for the clinical notes, one model can be trained using only the \textit{assessment} section of the notes, whilst another can be trained using the \textit{assessment}+\textit{subjective} sections.

Given an ensemble of diverse models, one may then combine them for a more robust ensemble system. A possible model combination method is weight averaging. Although weight averaging across training runs has shown success in image classification \cite{wortsman2022model}, weight averaging across different training runs is expected to work only when individual runs operate in similar weight spaces. This limits the types of combinations for weight averaging to only models with the same architecture and the same input format. As a result, we focus instead on two other methods of combination: token-level ensembling and Minimum Bayes Risk decoding \cite{rosti-etal-2007-combining, rosti-etal-2007-improved}.



\subsubsection*{Token-level Ensemble}
Token-level ensemble (also known as product-of-expectations) is a technique to improve the performance of sequence-to-sequence models by combining predictions from multiple models at the token-level \cite{sennrich2015improving, freitag2017ensemble, malinin2021uncertainty, gec-ensemble-dist}. Let us consider $M$ different models, where we want to generate an output sequence, $\mathbf y = y_0, y_1, \hdots,$ from an input sequence $\mathbf x$. In the standard decoding setup, we can generate each token sequentially:
\begin{equation}
    p(\mathbf y|\mathbf x) = \prod_i p(y_i|\mathbf x, y_{<i})
\end{equation}
In token ensembling, each token's probability is the average probability of the individual models:
\begin{equation}
    p(y_i|\mathbf x, y_{<i}) = \frac{1}{M}\sum_m p_m(y_i|\mathbf x, y_{<i})
\end{equation}

\subsubsection*{Minimum Bayes Risk (MBR) Decoding}
Given all possible output sequences, $\mathcal Y$, standard decoding (inference) strategies such as beam search are used to select the sequence with the greatest likelihood:
\begin{equation}
    \mathbf{y}^* = \argmax_{{\mathbf{y}}\in\mathcal Y}\left\{ p({\mathbf{y}}|\mathbf{x})\right\}
\end{equation}
However, the above method is not well aligned with the final reward metric, $\mathcal R$, used to assess the quality of samples (e.g. ROUGE-L). Following MBR decoding \cite{kumar-byrne-2004-minimum,  consensus-decoding}, we can seek to select the most \textit{average} sample, $\mathbf{y}^*\in\mathcal Y$, as per our desired reward metric:
\begin{equation}
    \mathbf{y}^* = \argmax_{{\mathbf{y}}\in\mathcal Y}\left\{ \mathbb E_{p(\tilde{\mathbf{y}}|x)}[\mathcal R(\mathbf{y}, \tilde{\mathbf{y}})] \right\}
    \label{eq:mbr_expectation}
\end{equation}
where $\mathbf{y}^*$ is expected to be the most representative of all generated samples. 
In practice, with only access to $N$ sequences, $\mathcal Y = \{\mathbf{y}_1, \hdots, \mathbf{y}_N\}$, sourced from different model structures, there may not be available sensible or meaningful posterior distributions, $p(\mathbf y|\mathbf x)$ and hence we approximate the expectation as a simple average, where each observed output $\mathbf y$  is taken to be equiprobable:
\begin{equation}
    \mathbf{y}^* \approx \argmax_{\mathbf{y} \in \mathcal{Y}}\left\{\sum_{n=1}^N \mathcal R(\mathbf{y}, \tilde{\mathbf{y}}_n) \right\}
    \label{eq:mbr_approx}
\end{equation}
The selection of $\mathbf{y}^*$ can also be viewed as a method to automatically reject the anomalous samples and thus improve overall performance. Previous work showed that MBR decoding improves machine translation \cite{rosti-etal-2007-combining, rosti-etal-2007-improved}, and the highest evaluation score is obtained when $\mathcal{R}$ matches the evaluation metric \cite{freitag-etal-2022-high}. Thus, we use ROUGE-L as the reward metric $\mathcal{R}$.

We note that MBR decoding is applied at inference time, and it can be applied to any set of models regardless of their architectures or training techniques, but it is expected to be effective when there is diversity in the models' outputs.


\subsubsection*{Hierarchical Ensembling}
Finally, hierarchical ensembling is a method that aims to combine the above two approaches. Multiple output sequences, $\mathcal Y$, can be generated by performing token ensembling over different sets of individual models. Subsequently, MBR decoding can be used to select the most representative sample from these different output sequences to give a single output sequence, $\mathbf y^*$. This hierarchical structure is depicted in Figure \ref{fig:diagram}.

\section{Experiments}
\subsection{Experimental Setup}
\noindent\textbf{Data.} Training data consists of 765 progress notes along with output medical summaries, which were sourced from MIMIC-III. Due to the small amount of data available, systems were initially evaluated using 5-fold cross-validation. The test (held-out competition) data consists of 237 progress notes, where for the competition evaluation we submitted generated summaries onto an online platform where the ROUGE-L was calculated. ROUGE-1, ROUGE-2 and ROUGE-L \cite{lin-2004-rouge} were all computed during cross-fold validation.



The medical reports have three fields available: `assessment' \texttt{\{A\}}, `objective' \texttt{\{O\}}, and `subjective' \texttt{\{S\}}, with word statistics shown in Table \ref{tab:data_stats}. We also consider different permutations by concatenating fields, separated by special tokens. \\

\begin{table}[!ht]
    \centering
    \small
    \setlength\extrarowheight{-2pt}
    \begin{tabular}{lcccc}
        \toprule
        Field      &\texttt{\{O\}}  &\texttt{\{S\}}  &\texttt{\{A\}} &Summary \\
        \midrule
        \#words &304.7\tiny{$\pm$83.4} &85.5\tiny{$\pm$54.8} &33.7\tiny{$\pm$17.1} &10.5\tiny{$\pm$7.5} \\
        \bottomrule
    \end{tabular}
    \caption{Medical report and summary statistics.}
    \label{tab:data_stats}
\end{table}


\noindent\textbf{Models.}
For abstractive summarization, we consider T5 and Clinical-T5 as the backbone. Clinical-T5\footnote{https://huggingface.co/xyla/Clinical-T5-Large.} was initialized from scratch and pre-trained on the union of MIMIC-III and MIMIC-IV databases \cite{lehman2023we}. The models are downloaded through HuggingFace; we finetune models with teacher forcing on our training data and use beam search during inference. More details about training and inference are provided in Appendix \ref{appendix:more_details}.



\subsection{Baseline Selection}
We start our investigation by comparing zero-shot, extractive, and abstractive summarization methods. Note that we provide the results of zero-shot summarization based on open-sourced large language models in Appendix \ref{appendix:zeroshot}.
\subsubsection*{Extractive Summarization} 

To obtain an \textit{empirical} upper bound, we compute ROUGE-1 between each input sentence against its ground-truth summary. The input sentences are ordered by the sections \texttt{\{A\}}, \texttt{\{S\}}, \texttt{\{O\}}. 

We consider two oracle options: (1) \textit{All-overlap}, which concatenates all input sentences where ROUGE-1 recall is positive to the generated summary; and (2) \textit{Greedy-best}, which uses a greedy algorithm to obtain extractive sentences similar to \citet{nallapati2017summarunner}. This greedy-best method iteratively adds sentences one at a time to the generated summary, where the added sentence is the one which yields the highest ROUGE-1 (F1) score. This process is repeated until the ROUGE-1 (F1) of the generated summary does not improve. Our results in Table \ref{tab:cross_valiation_results} show that even the oracle (greedy-best) approach achieves lower scores than fine-tuned T5 models (in Table \ref{tab:backbone-t5}).


\begin{table}[!ht]
  \centering
  \small
  \begin{tabular}{l|ccc}
    \toprule
    Method     &R1 &R2 &RL \\
    \midrule
    Oracle (All-overlap)  &20.37 &6.41  &15.44 \\
    Oracle (Greedy-best)  &29.14 &11.09 &22.74 \\
    \bottomrule
  \end{tabular}
  \caption{Empirical upper bounds of extractive summarization methods on the training data.}
  \label{tab:cross_valiation_results}
\end{table}

\subsubsection*{Abstractive Summarization}
Our first experiment is to determine the best transformer backbone for abstractive summarization. Table \ref{tab:backbone-t5} shows T5 performance when fine-tuned using the assessment field only. We find that T5-Large significantly outperforms T5-Small, but that performance does not further scale with size with T5-Large and T5-XL performing similarly. 

\begin{table}[!ht]
    \centering
    \small
    \begin{tabular}{rcccc}
        \toprule
        Pre-Trained Model & R1 & R2 & RL \\ 
        \midrule
        T5-Base          & 26.77      & 10.33      & 24.82 \\
        T5-Large         & 29.56      & 12.01      & 27.88 \\
        T5-XL            & 29.46      & 12.72      & 27.58 \\
        \midrule
        Clinical-T5-Base     & 26.62      & 12.11     & 24.96 \\
        Clinical-T5-Large    & \bf{32.22} & \bf{14.30} & \bf{30.15} \\
        \bottomrule
    \end{tabular}
    \caption{ROUGE-scores of various T5/Clinical-T5 models on cross-validation, with inputs being \texttt{\{A\}} only.}
    \label{tab:backbone-t5}
\end{table}

We further find that domain adaptation can lead to an additional boost, with Clinical-T5-Large showing ROUGE scores 2\% higher than T5-Large. Domain adaptation, however, was not helpful for our low-capacity model, with Clinical-T5-Base performing similarly to T5-Base. We, therefore, use \textbf{Clinical-T5-Large}, the best-performing system during cross-validation, as our backbone transformer for all further systems.

The next experiments consider which input fields are most useful for generating the summary. Table \ref{tab:input_perm} shows that the assessment field, \texttt{\{A\}}, contains the key information for the patient's problem  summary with ROUGE scores below 20 when any other field is used alone. We further observe better performance when \texttt{\{A\}} is augmented with \texttt{\{S\}}.

\begin{table}[!ht]
    \centering
    \small
    \begin{tabular}{ccccc}
        \toprule
        Inputs            & R1     & R2      & RL  \\ 
        \midrule
        $\emptyset$                & \;9.61 & \,2.93 & \,9.24 \\
        \texttt{\{O\}}             & 14.09   & 4.99    & 13.27 \\
        \texttt{\{S\}}             & 17.96   & 7.00    & 16.81 \\
        \texttt{\{A\}}             & 32.22   & 14.30   & 30.02 \\
        \midrule
        \texttt{\{A\}}+\texttt{\{S\}}       & 33.46   & 15.07    & 31.03\\
        \bottomrule
    \end{tabular}
    \caption{Comparison of Clinical-T5-Large performance when using different inputs on training data (using cross-validation). The empty input baseline $\emptyset$ is trained to generate summaries without the input report.}
    \label{tab:input_perm}
\end{table}

\subsection{Ensemble Methods}
\label{section:hesm}

To maximize the performance, we apply ensemble methods to fine-tuned Clinical-T5-Large models. For the simplicity of notation, we use $\theta_{\tt A}$ and $\theta_{\tt AS}$ to denote the system where the input is \texttt{\{A\}} and \texttt{\{A\}}+\texttt{\{S\}}, respectively. We train nine $\theta_{\tt A}$ individual models where all models are initialized from Clinical-T5-Large weights, and have the same training hyperparameters except random seeds for data batching. In Table \ref{tab:ensemble_Amodel_results}, we compare different methods for combining nine $\theta_{\tt A}$ models and the results show that: 1) weight averaging results in a slightly worse system; 2) both token-level ensemble and MBR decoding yield better performance than single models. In addition, we observe a similar trend when combining $\theta_{{\tt AS}}$ models as shown in Table \ref{tab:ensemble_ASmodel_results}.

\begin{table}[!ht]
    \centering
    \small
    \begin{tabular}{c|lll}
        \toprule
        \multirow{2}{*}{Method}   & \multicolumn{3}{c}{ROUGE-L} \\
        & F$_1$ & {Prec} & {Rec} \\
        \midrule
        Individual & 29.84{\tiny$\pm$0.69}  & 40.11{\tiny$\pm$1.40} & 27.68{\tiny$\pm$0.76} \\  
        Weight Avg.     & 29.39 & 39.68 & 27.50 \\
        Tok. Ensemble   & 30.50 & 41.09 & 28.37 \\ 
        MBR Decoding    & 30.72 & 40.96 & 28.91 \\
        \bottomrule
    \end{tabular}
    \caption{ROUGE-L on the test data. This table compares combination methods of nine $\theta_{{\tt A}}$ models.}
    \label{tab:ensemble_Amodel_results}
\end{table}

\begin{table}[!ht]
    \centering
    \small
    \begin{tabular}{c|lll}
        \toprule
        \multirow{2}{*}{Method}   & \multicolumn{3}{c}{ROUGE-L} \\
        & {F$_1$} & {Prec} & {Rec} \\
        \midrule
        Individual          &29.44{\tiny$\pm$0.45}  &37.57{\tiny$\pm$0.69} &28.33{\tiny$\pm$0.65} \\   
        Weight Avg.     &28.00  &35.65  &27.16 \\
        Tok. Ensemble   &30.04	&36.35	&29.96 \\ 
        MBR Decoding    &30.30  &38.39  &28.92\\
        \bottomrule
    \end{tabular}
    \caption{ROUGE-L on the test data. This table compares combination methods of nine $\theta_{{\tt AS}}$ models.}
    \label{tab:ensemble_ASmodel_results}
\end{table}


\noindent \textbf{Hierarchical Ensemble.} We explore combining $\theta_{\tt A}$ and $\theta_{\tt AS}$ models in a token-level ensemble followed by MBR decoding to form a hierarchical ensemble. Based on nine $\theta_{\tt A}$ models and nine $\theta_{\tt AS}$ models, Table \ref{tab:hierarchical_ensemble} provides the results of hierarchical combination in different setups.

The first block shows the performance when combining one $\theta_{\tt A}$ and one $\theta_{\tt AS}$ in a token-level ensemble, followed by an MBR combination stage over 9 of these ensembles. Similarly, the second block shows the performance when combining three $\theta_{\tt A}$ and three $\theta_{\tt AS}$ each in a token-level ensemble fashion followed by an MBR decoding stage over 3 of these ensembles. 

\begin{table}[!ht]
    \tabcolsep=0.11cm
    \centering
    \small
    \begin{tabular}{c|cc|lll}
        \toprule
        \multirow{2}{*}{Name}& \multicolumn{2}{c|}{Ensemble} & \multicolumn{3}{c}{ROUGE-L} \\
        & Token & MBR & {F$_1$} & {Prec} & {Rec} \\
        \midrule
        $\theta_{\tt A}$+$\theta_{\tt AS}$ & (1, 1) & \xmark & 31.17\tiny{$\pm$0.67} & 39.51\tiny{$\pm$1.30} & 29.66\tiny{$\pm$1.02} \\
        HESM & (1, 1) & 9 & 32.31 & 41.16 & 30.16 \\ 
        \midrule
        $\theta_{\tt A}$+$\theta_{\tt AS}$ & (3, 3) & \xmark & 31.50\tiny{$\pm$0.42} & 39.74\tiny{$\pm$0.79} & 29.97\tiny{$\pm$0.57} \\ 
        \multirow{2}{*}{HESM} & (3, 3) & 3 & 31.87 & 39.63 & 30.24 \\ 
        & (3, 3) & 9 & 31.88 & 40.07 & 30.17 \\ 
        \bottomrule
    \end{tabular}
    \caption{ROUGE-L of HESM on the test data. ($a$, $b$) denotes token-level ensemble consisting of $a\theta_{\tt A}$ models and $b\theta_{\tt AS}$ models. MBR=$c$ denotes the outputs of $c$ token-level ensembles combined using MBR decoding. For HESM(3,3) w/ MBR=3, ensembles with non-overlap members are chosen.}
    \label{tab:hierarchical_ensemble}
\end{table}

\subsection{Evaluation System}
This section discusses the specific nature of our HESM systems submitted to the shared task. Given the flexibility in an MBR combination, members of HESMs are not limited to token-level ensembles. Hence, during the competition we made use of previously submitted systems to build the final HESM. As a result, our HESM consists of six systems: best-performing $\theta_{{\tt A}}$; weight averaging of 3$\theta_{{\tt 
 A}}$; token-level ensemble of 3$\theta_{{\tt A}}$ with $\mathcal{L}_\text{RL}$; 2$\times$token-level ensemble of 9$\theta_{{\tt A}}$; and token-level ensemble of 9$\theta_{{\tt AS}}$. The results of the HESM's members are provided in Table \ref{tab:member_of_hesm24} in the appendix.

We further consider combining this HESM with the token-level ensembles of 3$\theta_{{\tt A}}$+3$\theta_{{\tt AS}}$ investigated in Section \ref{section:hesm}. The first ensemble (v1) is obtained by selecting three $\theta_{{\tt A}}$ and three $\theta_{{\tt AS}}$ (out of the nine $\theta_{{\tt A}}$ and nine $\theta_{{\tt AS}}$) with the lowest cross-entropy training losses. The second ensemble (v2) is obtained by training variants of the three $\theta_{{\tt A}}$ and three $\theta_{{\tt AS}}$ in the first ensemble using different hyperparameters to increase diversity. The results of these two token-level ensembles are reported in Table \ref{tab:hesm_competition}.

Ultimately, we combine the above HESM with these two ensembles using MBR decoding, and this combined system can be viewed as a higher level of HESM as it consists of HESM as a member of the MBR combination. This final combination sets the state-of-the-art performance of the task, achieving the ROUGE-L score of 32.77.

\begin{table}[!ht]
    \centering
    \small
    \begin{tabular}{c|ccc}
    \toprule
    \multirow{2}{*}{System} &\multicolumn{3}{c}{ROUGE-L} \\
         &F$_1$      &Prec &Rec \\
    \midrule
    HESM            &31.86	&43.52	&28.90 \\  
    TokEns(3$\theta_{\tt A}$+3$\theta_{\tt AS}$)-v1    &32.03	&41.01	&30.16  \\ 
    TokEns(3$\theta_{\tt A}$+3$\theta_{\tt AS}$)-v2    &32.19	&39.59	&30.88  \\ 
    \midrule
    + MBR Combination$^\dagger$   &\textbf{32.77} &41.69	&30.51 \\  
    \bottomrule
    \end{tabular}
    \caption{ROUGE-L on the test data. $^\dagger$This system attains the top position on the shared task leaderboard. }
    \label{tab:hesm_competition}
\end{table}
\section{Conclusions}
In low-resource and medical-domain summarization, our work has demonstrated that abstractive summarization outperforms extractive and zero-shot methods. Furthermore, both token-level ensemble and MBR decoding improve the overall performance. Our HESM, which utilizes both ensembling techniques, achieves state-of-the-art performance with the highest ROUGE-L score in the BioNLP 2023's shared task 1A leaderboard.


\section{Limitations}

The limitations of this work are mainly that there is a small amount of data available for inference to test the models. ROUGE-L is used as an assessment metric and n-gram overlap metrics are notably not optimal for abstractive summarization assessment \cite{bertscore_paper, deutsch2022evaluation}.

\section{Ethics Statement}

The study used de-identified health data to develop a system that overcomes biases in medical decision-making. However, social biases in language models need to be addressed to ensure fairness in model training. Therefore, before deploying any pre-trained language model, fairness audits are necessary to ensure an ethical and trustworthy model for all stakeholders.
Note, doctors should not rely on automated summarization systems for diagnoses in the interest of patient care.

\section*{Acknowledgements}
This paper reports on research supported by Cambridge University Press \& Assessment (CUP\&A), a department of The Chancellor, Masters, and Scholars of the University of Cambridge. This research is further supported by the EPSRC (The Engineering and Physical Sciences Research Council) Doctoral Training Partnership (DTP) PhD studentship, the Cambridge International \&
St John’s College scholarship, and the Gates Cambridge Scholarship.

\bibliography{anthology,custom}
\bibliographystyle{acl_natbib}

\appendix
\section{More details about experiments}
\label{appendix:more_details}
\subsection{Inference Hyperparameters} 
num\_beams = 4, length\_penalty = 0.6, min\_length = 5, max\_length = 256, no\_repeat\_ngram\_size = 4.
\subsection{RL training}
\label{appendix:rl_training}
We follow \citet{paulus2018a} in using reinforcement learning (RL) based loss:
\begin{equation}
    \mathcal{L}_{\text{RL}} = (\mathcal{R}(\bar{\mathbf{y}},\mathbf{y}) - \mathcal{R}(\hat{\mathbf{y}},\mathbf{y})) \log P(\hat{\mathbf{y}}|\mathbf{x})
\end{equation}
where $\bar{\mathbf{y}}$ is the sequence obtained by greedy search, $\hat{\mathbf{y}}$ is the sequence obtained by sampling, and $\hat{\mathbf{y}}$ is the ground-truth sequence. To improve the stability of training, we initialize the model using the weights from the maximum likelihood training (cross-entropy loss), and we use a combined loss: $\mathcal{L} = \gamma \mathcal{L}_{\text{RL}} + (1-\gamma)\mathcal{L}_{\text{ML}}$ where $\gamma$ = 0.9 and $\mathcal{L}_{\text{ML}}$ is the standard cross entropy loss. The results are provided in Table \ref{tab:member_of_hesm24}, showing that a marginal gain can be achieved from RL training. 

\section{Additional Results}

\subsection{Zero-shot Summarization}
\label{appendix:zeroshot}
Since state-of-the-art LLMs such as GPT-3 or ChatGPT are only available via API services, using them would violate the MIMIC data use agreement. Instead, we use open-source LLMs.
We use the following text input to large language models (LLMs),
\begin{verbatim}
    {prompt}:
    {clinical_note}
\end{verbatim}
where \texttt{clinical\_note} is the \textit{assessment} section, and we consider two prompts:
    \begin{itemize}
        \item P1: \textit{Summarize these clinical notes.}
        \item P2: \textit{Give a one or two word summary for these clinical notes.}
    \end{itemize}
We use open-source LLMs including OPT-IML \cite{iyer2022opt}, GPT-J \cite{gpt-j}, and GPT-NeoX \cite{gpt-neox-20b}. We provide the results on training data in Table \ref{tab:zeroshot_results}. The poor performance could be attributed to the small size of LLMs, and larger systems such as GPT-3 or ChatGPT could potentially perform much better.

\begin{table}[!ht]
    \centering
    \small
    \begin{tabular}{l|c|ccc}
    \toprule
        LLM &prompt &R1 &R2 &RL  \\ 
        \midrule
        OPT-IML-1.3B &P1 &4.97  &0.80  &4.37 \\
                     &P2 &4.46  &0.65  &4.05 \\
        \midrule
        OPT-IML-30B  &P1  &2.76  &0.44  &2.51 \\
                     &P2  &2.07  &0.36  &1.96 \\              
        \midrule
        GPT-J-6B     &P1  &4.13  &0.58  &3.68 \\
                     &P2  &4.66  &0.73  &4.29 \\        
        \midrule
        GPT-NeoX-20B &P1  &2.41  &0.38  &2.21 \\
                     &P2  &3.04  &0.59  &2.79 \\    
         \bottomrule
    \end{tabular}
    \caption{Zero-shot Summarization performance on training data of LLMs with different user prompts.}
    \label{tab:zeroshot_results}
\end{table}

\subsection{More Analysis}
Table \ref{tab:input_perm2} shows our post-evaluation studies on the performance using different input fields, and the results suggest that it is possible to improve the performance further by using \texttt{\{A\}}+\texttt{\{S\}}+\texttt{\{O\}} in addition to \texttt{\{A\}} and \texttt{\{A\}}+\texttt{\{S\}} as the input.
\begin{table}[!ht]
    \centering
    \small
    \begin{tabular}{lcccc}
        \toprule
        Inputs            & R1     & R2      & RL  \\ 
        \midrule
        \texttt{\{A\}}             & 32.22   & 14.30    & 30.02 \\
        \midrule
        \texttt{\{A\}}+\texttt{\{O\}}       & 32.50   & 13.81    & 30.31 \\
        \texttt{\{A\}}+\texttt{\{S\}}       & 33.46   & 15.07    & 31.03\\
        \texttt{\{A\}}+\texttt{\{S\}}+\texttt{\{O\}} & 33.80   & 15.38    & 31.28 \\
        \bottomrule
    \end{tabular}
    \caption{Comparison of Clinical-T5-large performance when using different inputs on training data (using cross-validation).}
    \label{tab:input_perm2}
\end{table}

\subsection{Submitted Systems}
\label{appendix:more_results}
In Table \ref{tab:member_of_hesm24}, we present other approaches that were submitted to the shared task, including model weight averaging, and RL-based training. These models also formed components of the final HESM model submitted.
\begin{table}[!ht]
  \centering
  \small
  \begin{tabular}{l|ccc}
    \toprule
    \multirow{2}{*}{System}         &\multicolumn{3}{c}{ROUGE-L} \\
    &F$_1$ &Prec &Rec \\
    \midrule
    Weight Avg. of 3$\theta_{{\tt A}}$                 &30.26	&42.51	&27.31  \\ 
    TokEns 3$\theta_{{\tt A}}$ w/ $\mathcal{L}_{\text{RL}}^\dagger$  &30.40	&43.78	&27.10  \\ 
    Best-performing $\theta_{{\tt A}}$ &30.56	&39.97	&28.95 \\
    TokEns 9$\theta_{{\tt A}}$-v1 &30.74	&42.14	&27.93 \\ 
    TokEns 9$\theta_{{\tt A}}$-v2 &30.50	&41.09	&28.37 \\ 
    TokEns 9$\theta_{{\tt AS}}$   &30.04	&36.35	&29.96 \\ 
    \bottomrule
  \end{tabular}
  \caption{ROUGE-L scores on test data of the members of HESM. $^\dagger\mathcal{L}_{\text{RL}}$ is described in Appendix \ref{appendix:rl_training}.}
  \label{tab:member_of_hesm24}
\end{table}

\section{Post-evaluation Ablation Study}
The results in Table \ref{tab:hesm_competition} found that a higher-level Hierarchical Ensemble (HESM) model had the best performance. This model performs MBR decoding over the output from an existing shallower HESM model (we will refer to as HESM-shallow) and 2$\times$token-level ensemble of 3$\theta_{{\tt A}}$+3$\theta_{{\tt AS}}$. Table \ref{tab:testset_mbr} explores the impact on performance with unpacking the token level-ensemble systems and performing MBR decoding over all individual systems. The system labelled with \textit{unpack-1} performs MBR decoding over the 6 ensemble systems that form HESM-shallow and the 12 individual systems used to make the two token level ensemble systems; i.e. MBR decoding is performed over \textbf{18} system output sequences. The \textit{unpack-2} system considers further unpacking the 6 ensemble systems used for HESM-shallow, such that MBR decoding is now performed over a total of \textbf{34} unique individual systems.

\begin{table}[!ht]
  \centering
  \small
  \begin{tabular}{l|ccc}
    \toprule
    \multirow{2}{*}{System}         &\multicolumn{3}{c}{ROUGE-L} \\
    &F$_1$ &Prec &Rec \\
    \midrule
    HESM (final)  &32.77 &41.69	&30.51\\ 
    \midrule
    HESM-unpack-1   &32.38 & 42.98 & 29.91\\
    HESM-unpack-2   &32.26 & 43.18 & 29.62 \\
    \bottomrule
  \end{tabular}
  \caption{ROUGE-L scores on the test data. This table considers the impact of performing MBR decoding on the individual systems after unpacking token-level ensemble systems used as components for the final higher-level HESM model submitted in the competition in Table \ref{tab:hesm_competition}.}
  \label{tab:testset_mbr}
\end{table}

\end{document}